\documentclass[10pt,twocolumn,letterpaper]{article}

\usepackage{iccv}
\usepackage{times}
\usepackage{epsfig}
\usepackage{graphicx}
\usepackage{amsmath}
\usepackage{amssymb}

\usepackage{float}
\usepackage{colortbl}
\usepackage{multirow, booktabs}
\usepackage{siunitx}
\usepackage{fixltx2e}
\usepackage{setspace}
\usepackage{booktabs}
\usepackage{array}
\usepackage{tabularx}
\usepackage{calc}
\usepackage{makecell}
\usepackage{comment}
\usepackage{dsfont}
\usepackage{multirow}

\usepackage{subcaption}

\newcommand{\calF}{\mathcal{F}}
\newcommand{\calP}{\mathcal{P}}
\newcommand{\vn}{\mathbf{n}}
\newcommand{\vx}{\mathbf{x}}

\def\x{\times}
\newcommand{\PAR}[1]{\vskip4pt \noindent{\bf #1~}}

\newcommand{\OURS}{UniPlane}

\usepackage[breaklinks=true,bookmarks=false,hidelinks]{hyperref}

\iccvfinalcopy 

\def\iccvPaperID{6438} 

\begin{document}

\title{UniPlane: Unified Plane Detection and Reconstruction\\from Posed Monocular Videos}


\author{
Yuzhong Huang$^{1,2,*}$~~~~~Chen Liu$^{2}$~~~~~Ji Hou$^{2}$~~~~~Ke Huo$^{2}$~~~~~Shiyu Dong$^{2}$~~~~~Fred Morstatter$^{1}$\vspace{0.2cm}\\
$^{1}$University of Southern California~~~~~$^{2}$Meta Reality Labs
}

\twocolumn[{%
	\renewcommand\twocolumn[1][]{#1}%
	\maketitle
	\begin{center}
 
	\includegraphics[width=\linewidth]{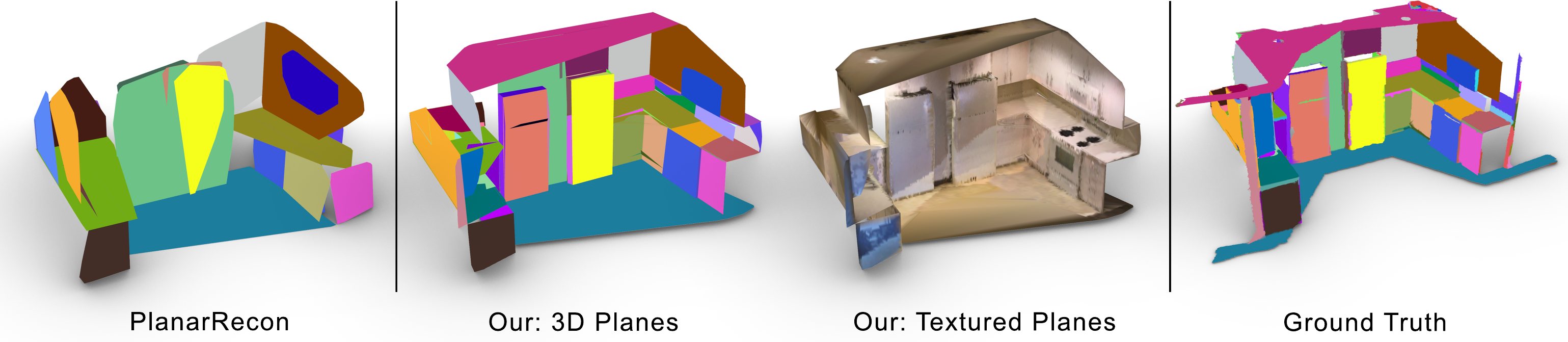}
		\captionof{figure}{
                \textbf{Comparison between \OURS{} and PlanarRecon}. Left: predictions from our baseline PlanarReccon. Middle: reconstructions from \OURS{}. Right: ground-truth plane reconstruction. Each color represents a plane instance. Textured planes are learned with rendering loss. Our model is able to accurately detect more planes improving both recall and precision. 
		}
		\label{fig:teaser}
	\end{center}
}]

\maketitle

{\let\thefootnote\relax\footnote{{
{$^{*}$} Work done while Yuzhong was an intern at Meta Reality Labs
 }}}
 
\begin{abstract}
We present \OURS{}, a novel method that unifies plane detection and reconstruction from posed monocular videos. Unlike existing methods that detect planes from local observations and associate them across the video for the final reconstruction, UniPlane unifies both the detection and the reconstruction tasks in a single network, which allows us to directly optimize final reconstruction quality and fully leverage temporal information. Specifically, we build a Transformers-based deep neural network that jointly constructs a 3D feature volume for the environment and estimates a set of per-plane embeddings as queries. \OURS{} directly reconstructs the 3D planes by taking dot products between voxel embeddings and the plane embeddings followed by binary thresholding.
Extensive experiments on real-world datasets demonstrate that \OURS{} outperforms state-of-the-art methods in both plane detection and reconstruction tasks, achieving +4.6 in F-score in geometry as well as consistent improvements in other geometry and segmentation metrics.


\end{abstract}
\section{Introduction}
\label{sec:intro}
3D reconstruction is one of the key topics in computer vision, which is fundamental to many high-level 3D perception and scene understanding tasks~\cite{hou20193d, hou2020revealnet, nie2021rfd}. 3D reconstruction is the process of reconstructing the surrounding environment from sensor input, which may include color images, depth images, and camera poses. The resulting reconstructed scene can take the form of a point cloud, mesh, or implicit surface. In this paper, we aim for implicit surface reconstruction since applications, like augmented reality, virtual reality, and autonomous driving, require a compact and explainable representation.

Considerable effort \cite{schops3DModelingGo2015,yangRealtimeMonocularDense2017} has been devoted to 3D reconstruction using depth data. However, even with dedicated depth sensors, depth data remain unreliable and are affected by changing light conditions and surface materials \cite{newcombeKinectFusionRealTimeDense2011}. Moreover, depth sensors are often less computation-efficient, and are infeasible to be mounted on many mobile devices, such standalone VR/AR headsets and many non-flagship phones. Therefore, visual-only approaches have gained a lot of interest. Currently, the typical pipeline involves first estimating depth images from color images \cite{mvdepthnet, Liu_2019, 9008539, 10.1145/3272127.3275041, dahnert2021panoptic} and then using depth fusion to reconstruct 3D scene geometry. However, these two-stage approaches are unable to jointly optimize estimated depths whose quality is often the performance bottleneck.

Recently, NeuralRecon~\cite{sun2021neucon} proposed an approach that avoids estimating view-dependent 2D depth maps by projecting 2D image features into a 3D voxel grid, allowing for tracking and fusion of multiple input frames. PlanarRecon~\cite{xie2022planarrecon}extends the framework to reconstruct explicit planar surfaces and demonstrates that the dedicated explicit planar surface reconstruction framework achieves much better performance than adding a separate surface extraction module (e.g., RANSAC plane fitting module) on top of NeuralRecon.
However, PlanarRecon still has several limitations. First, it only supervises individual voxels to predict per-voxel plane attributes, and clusters voxels into plane instances using mean-shift. The lack of instance mask supervision on the clustering results often leads to inaccurate boundaries and problematic segmentation (e.g., over-segmentation). The second limitation is that it could only handle a video fragment (i.e., 9 frames) in the detection stage, and a tracking and fusion module is needed to merge results across video fragments. Though the tracking module utilizes learned features as one metric, it still heavily relies on other hand-crafted heuristics. Merging failures in the tracking and fusion module lead to duplicate planes even with good single-fragment results.

To address these issues, our \OURS{} deploys a Transformers-based deep network to jointly construct a 3D embedding volume for the environment and estimate a set of per-plane embeddings as queries following the idea of~\cite{cheng2021per}. During training, the dot products between  voxel embeddings and plane embeddings are supervised using ground truth segmentation. During inference, we threshold dot products to reconstruct plane masks. We can directly model 3D planes using reconstructed plane masks and predicted plane parameters. In addition, we exploit the sparse nature of the 3D feature volume by attending to the high-occupancy region. Furthermore, the reconstructed surfaces are refined using a rendering loss with respect to input images.

Similar to PlanarRecon, our method takes a monocular video as input, operating on a 3D feature volume fused by a 2D image sequence. Different from PlanarRecon, our method queries per-object embedding vector over the entire scene. To effectively query the scene, we propose to leverage a sparse attention. To further improve the reconstruction quality, we utilize view consistency to eliminate the null space and refine the generated surface with rendering loss, in which our approach predicts a set of 3D surfaces whose renderings match the original 2D images.

The surface-based representation is a more effective model for representing real-world environments. It is a sparse representation, which avoids wasting resources on empty space. Common objects such as floors, walls, ceilings, beds, and desks can be efficiently modeled as surfaces with height fields. This approach leverages prior knowledge that 3D objects are sparsely distributed throughout the environment, and that each object occupies a continuous region in space.

To summarize, our contributions are three-folds: \begin{enumerate}
    \item With the differentiable instance segmentation network, \OURS{} is able to learn more accurate plane boundaries using segmentation supervision.
    \item The learned per-plane embedding vectors enable feature-based plane tracking across the entire video.
    \item \OURS{} further improves the surface reconstruction quality by filtering null space voxels based on view consistency and enforcing a rendering loss.
\end{enumerate}

\section{Related work}
\label{sec:related_work}
\subsection{Single-view plane reconstruction}
For plane detection from single-images, most traditional methods, \cite{silberman2012indoor,deng2017unsupervised}, rely on depth input. Recently, learning-based approaches, \cite{yuSingleImagePiecewisePlanar2019,liu2018planenet,liu2019planercnn,yang2018recovering,tanPlaneTRStructureGuidedTransformers2021}, formulate the task as an instance segmentation problem and train deep networks to jointly estimate the plane instance segmentation and per-instance plane parameters. These methods are able to achieve high-quality plane instance segmentation but the predicted 3D geometries are less accurate due to the ill-posed nature of the task.
Following works~\cite{qian2020learning,sun2021indoor}, we improve the reconstruction accuracy by predicting and enforcing the relationships between plane instances. Even then, the predicted geometries are not centimeter-accurate, making it extremely challenging to develop a robust multi-view tracking system upon these single-view detection networks.

\subsection{Multi-view plane reconstruction}
There has been extensive research on multi-view plane reconstruction with known camera poses. Early approaches~\cite{bartoli2007random,baillard1999automatic} first reconstruct a sparse set of 3D points or line features, which were then grouped together using certain heuristics. However, these methods heavily relied on hand-crafted features and were not robust to lighting changes or textureless regions. Other methods \cite{Furukawa09,sinha2009piecewise,gallup2010piecewise,gallup2010piecewise} approach the problem as an image segmentation task, where each pixel is assigned to one of the plane hypotheses using MRF formulation.

Jin et al.~\cite{Jin_2021_ICCV} extend the single-view learning approach \cite{liu2019planercnn}
to two views, and add optimization to maximum the consistency between two sets of planes reconstructed from each view. This pairwise optimization can be repeated to handle a few more images, but the process is time-consuming and it is unclear how to generalize it to more frames (e.g., the entire video).

Recently, PlanarRecon~\cite{xie2022planarrecon} builds a learning-based multi-view plane reconstruction system following the volumetric reconstruction approach of NeuralRecon ~\cite{sun2021neucon}. By learning planes directly from multi-view in an end-to-end manner, PlanarRecon outstands from its counterparts, especially for geometry accuracy. While PlanarRecon pushes the frontier of plane reconstruction, it has two limitations: 1) its clustering-based segmentation is often inaccurate due to the lack of instance-level supervision on the segmentation, 2) it produces plane duplicates when its tracking and fusion module fails to merge planes properly. To address these limitations, we train transformers to jointly learn per-instance plane features and segmentation masks (i.e., matching between plane features and voxel features). The learned plane features can also be used to track planes across video fragments without the need of hand-crafted heuristics.

\subsection{Learning-based tracking and reconstruction}
While deep networks are able to detect object instances from single images or video fragments and extract corresponding features, the instance association across frames is still challenging since object features are view-dependent due to lighting changes and partial observation. 

Many papers study image correspondence learning by developing feature-matching networks. Here, we discuss a few notable ones and refer to~\cite{ma2021image} for a comprehensive survey. SuperGlue designs a feature-matching network to estimate the correspondences between learned image features. The key contribution of the SuperGlue is to use attention mechanisms to focus on the most discriminative parts of the images and filter out irrelevant features during the matching process. TransMatcher~\cite{liao2021transmatcher} uses Transformers to apply the attention mechanisms. However, one cannot directly apply these approaches to address the problem of object instance tracking. Object tracking requires learned object features to be view-independent (e.g., similar features for two sides of the same chair) and discriminative (e.g., a chair's feature should be different from the others), and learning such features is not trivial. For this reason, even though PlanarRecon~\cite{xie2022planarrecon} deploys SuperGlue~\cite{sarlin20superglue} in their tracking and fusion module, it only uses the feature-based matching as one optional metric among many other hand-crafted matching metrics which are shown to be more important (e.g., Intersection-over-Union and plane-to-plane distance).

Several attempts have been made to tackle the object tracking problem. TrackFormer~\cite{meinhardt2022trackformer} deploys a Transformer to detect objects and learn the trajectories of the tracked objects in an end-to-end manner. However, TrackFormer focuses on the multi-object tracking setting where tracked objects have a smooth 2D trajectory on the image with a similar viewing angle. This generic image-based tracking approach is not not applicable to the object instance tracking problem where viewing angles change dramatically.

To utilize object priors as a rescue, ODAM performs object detection, association, and mapping jointly using a novel optimization framework. The key is to predict the object poses which are used as prior in the joint optimization. Wang et al.~\cite{wang2021multi} develop an end-to-end object reconstruction framework upon a volumetric feature representation constructed by a 3D volume transformer. However, these approaches focus on object reconstruction where object-specific priors like object bounding box and object-centric feature volume can be used. It is unclear how to extend them to reconstruct large environments. In this paper, we track and reconstruct large planar surfaces in large environments using a Transformer network that operates on a sparse volume.
\section{Methods}
This section explains how UniPlane constructs the 3D feature volume (Sec.~\ref{sec:volume}), detects planes differentiably from 3D feature volumes using Transformers (Sec.~\ref{sec:detection}), unifies plane reconstruction with detection (Sec.~\ref{sec:reconstruction}, and refines geometries with a rendering loss (Sec.~\ref{sec:rendering}.

\begin{figure}[t]
    \centering

    \includegraphics[width=\linewidth]{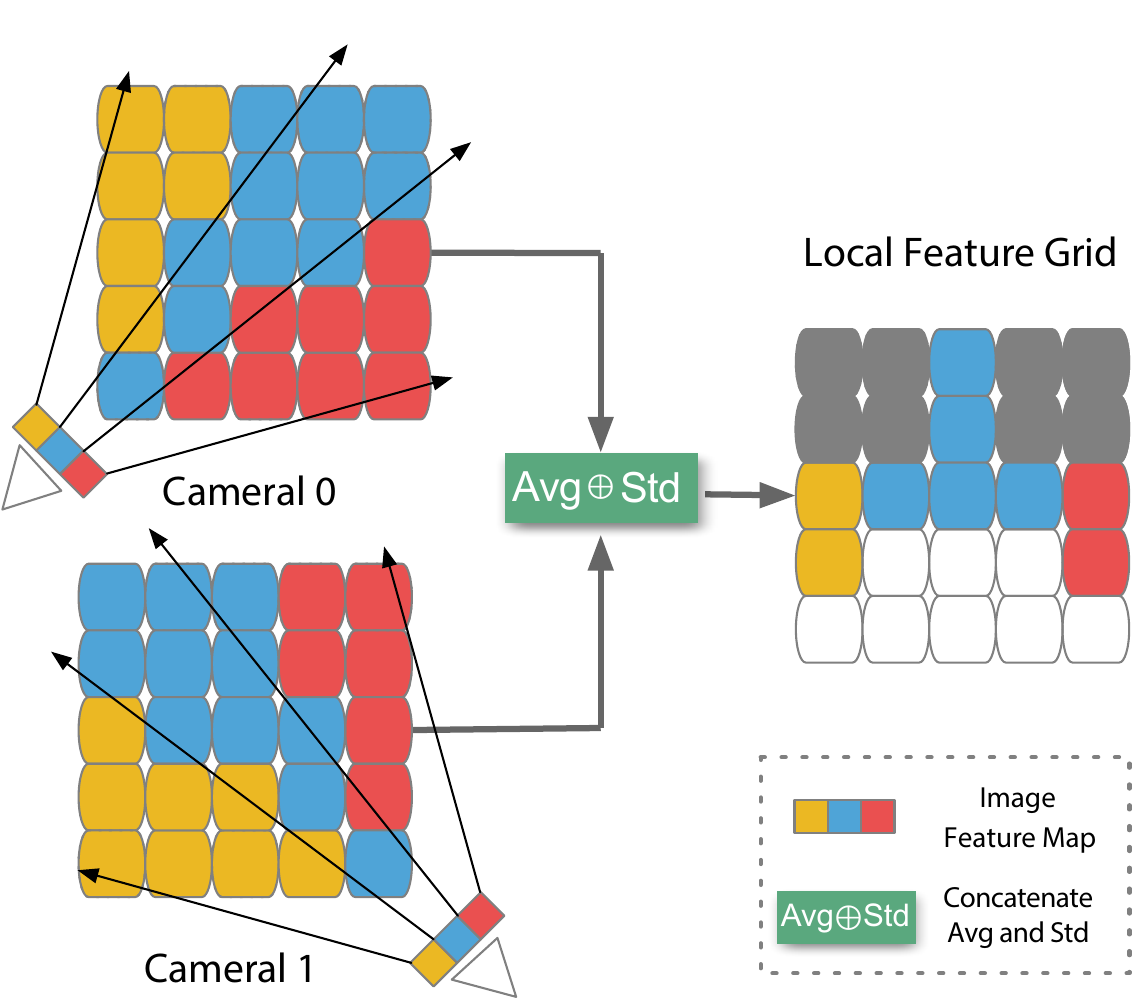}
       \caption{\textbf{2D toy example to illustrate view consistency}. Each 2D pixel will project visual features onto voxels accessible by a ray from it. Voxels receiving consistent visual features are occupied, marked by color on the right. Voxels receiving different visual features are unoccupied, and marked as white on the right. Voxels behind occupied voxels are occluded, and marked as gray on the right.}
       \label{fig:unproject}

\end{figure}

\begin{figure*}[h!]
    \centering
       \includegraphics[width=\linewidth]{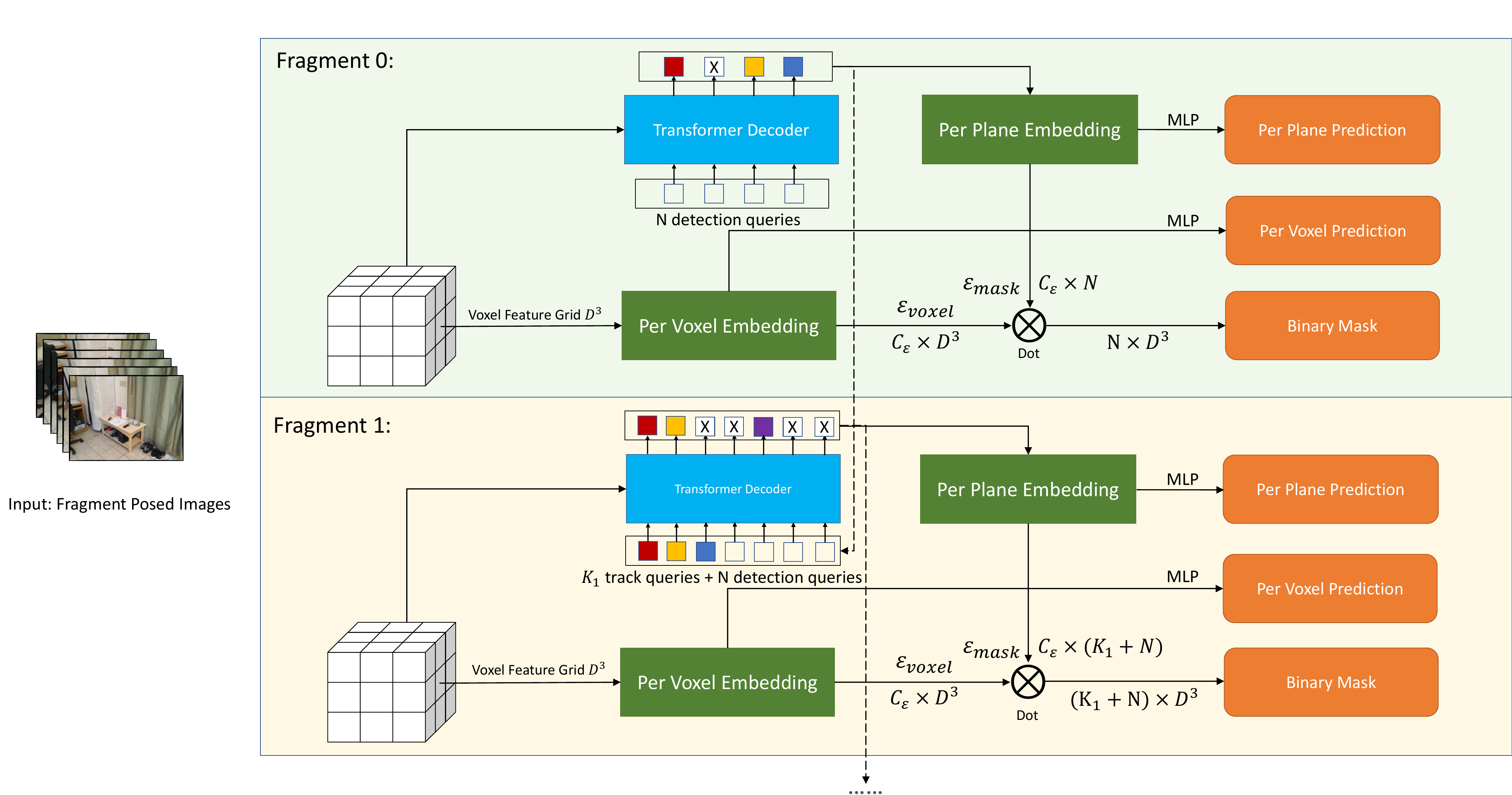}
       \caption{
           \textbf{Overall Architecture} From a sequence of posed images, we organize them into fragments. A voxel feature grid is constructed for each fragment. This per voxel feature is both used to make a per voxel prediction, and as key and value vector to a transformer decoder. The query vector to the transforms consists of both a sequence of learnable query vector, and plane embedding from previous fragment to track planes across fragments}
       \label{fig:arch}
       \vspace{0.5cm}
\end{figure*}




 \subsection{Feature volume construction}
 \label{sec:volume}
 
Following PlanarRecon~\cite{xie2022planarrecon}, we use a 2D backbone named MnasNet \cite{Tan_2019_CVPR} to extract features from 2D images and unproject image features to a shared 3D volume space. A voxel takes the average features from multiple pixels across different views. PlanarRecon~\cite{xie2022planarrecon} deploys sparse 3D convolution to exploit the sparsity of the feature volume by predicting which voxels are occupied and attending to those voxels.

However, the occupancy prediction of PlanarRecon is not effective to prune freespace voxels. For a freespace voxel, its feature mixes visual information corresponding to different surface points in different views (i.e., the actual surface point has the same image projection as the voxel but different depths). The mixed feature is often not discriminative enough for occupancy prediction. We improve the occupancy prediction by adding the standard deviation as the input feature. The intuition is that an occupied voxel on an actual surface should have similar features across different views, and thus its feature standard deviation should be small providing direct signals for occupancy prediction. This is similar to the view consistency idea widely used in the literature.

We further conduct experiments on adding the 2D feature standard deviation. Experiments show that adding the 2D feature standard deviation helps occupancy prediction and the final plane prediction.

\subsection{Transformers-based Plane Detection}
\label{sec:detection}
Following PlanarRecon~\cite{xie2022planarrecon}, we get per voxel embedding $\xi_{voxel} \in C_{\xi} \times D^3 $. Then by performing occupancy analysis, we get occupied voxel set $M$ and per-voxel features $\xi_{voxel} \in C_{\xi} \times M $, which will be used to regress the plane parameters, including the surface normal $\vn_m\in \mathbb{R}^3$ and plane offset $d_m\in \mathbb{R}^1$, plane center $c_m \in \mathbb{R}^3$ associated to the corresponding voxel.

As illustrated in Figure~\ref{fig:arch}, each voxel makes a prediction of the location and normal of the primitive it belongs to. Once we have geometric primitives for every single voxel (shifted voxels $\vx_m' \in \mathbb{R}^3$ and surface normals $\vn_m\in \mathbb{R}^3$), we group the voxels to form plane instances in the local fragment volume $\calF_i$. 

Unlike PlanarRecon~\cite{xie2022planarrecon} which uses k-means to perform voxel to plane assignment, we train a volume segmentation network that predicts instance segmentation differentiable following the idea of MaskFormer~\cite{cheng2021per}. The segmentation network avoids hyperparameter tuning, and enables direct segmentation supervision.

To construct the supervision, we find the matching $\sigma$ between the set of predictions $z$ and the set of $N^{\text{gt}}$ ground truth segments $z^\text{gt} = \{( c_{i}^\text{gt},  m_{i}^\text{gt}) | c_{i}^\text{gt} \in \{1, \dots, K\}, m_{i}^\text{gt} \in \{0, 1\}^{H \x W}\}_{i=1}^{N^{\text{gt}}}$ is required. Here $c_{i}^\text{gt}$ is the ground truth class of the $i^{\text{th}}$ ground truth segment.
Since the size of prediction set $|z| = N$ and ground truth set $|z^\text{gt}| = N^\text{gt}$ generally differ, we assume $N \ge N^{\text{gt}}$ and pad the set of ground truth labels  with ``no object'' tokens $\varnothing$ to allow one-to-one matching.

To train model parameters, given a matching, the main mask classification loss $\mathcal{L}_{\text{mask-cls}}$ is composed of a cross-entropy classification loss and a binary mask loss $\mathcal{L}_\text{mask}$ for each predicted segment:
\begin{multline}
    \label{eq:mask_cls}
    \mathcal{L}_{\text{mask-cls}}(z, z^\text{gt}) =\\ \sum\nolimits_{j=1}^N \left[-\log p_{\sigma(j)}( c_{j}^\text{gt}) + \mathds{1}_{c_{j}^\text{gt}\neq\varnothing} \mathcal{L}_{\text{mask}}(m_{\sigma(j)},  m_{j}^\text{gt})\right].
\end{multline}

The learning-based fully differentiable segmentation network achieves better segmentation quality and enables unified reconstruction as we discuss in the following section.

\subsection{Unifying Tracking with Reconstruction}
\label{sec:reconstruction}
Inspired by TrackFormer~\cite{meinhardt2022trackformer}, we unify plane tracking and reconstruction into the same volume segmentation network, by using per-plane features from previous fragment as tracking queries in the next fragment. (Fig.~\ref{fig:arch}).

In fragment $i$, detection query $n$ detected plane instance $\calP_i^n$. Then we use the same query $n$ as tracking query in fragment $i+1$. If it matches a plane instances $\calP_{i+1}^n$, then $\calP_i^n$ will be merged with $\calP_{i+1}^n$.

By integrating reconstruction and detection in the same unified framework, we eliminate the need for a separate heuristic-heavy tracking and fusion module. This end-to-end network outperforms the two-stage PlanarRecon~\ref{tab:scannet-3d}, having less parameters and is easier to train.

\subsection{Refine Planes using Differentiable Rendering}
\label{sec:rendering}

Each detected plane could be represented by (1) Plane center (2) Plane coefficient (normal, offset) (3) Plane boundary function $f(\theta) \rightarrow r$ (4) Plane color function $f(\theta, r) \rightarrow RGB$. Plane center and coefficients will be initialized according to predictions in the previous stage.

Plane boundary function $f_{boundary}(\theta)$ is a MLP initialized according to the boundary of voxels of this plane. We assume the plane is convex, and using a distance function from plane center to represent the plane, allow planes in different shape to be represented in a uniform way, and enable refinement of plane shape using differentiable rendering. Plane color function $f_{color}(\theta, r)$ is a MLP need to be learned. For simplification, we assume the planes are opaque and their appearance is view independent.

Given a camera with an associated pose, we can render a predicted image. For each pixel in the image we can generate a ray according to camera intrinsic and extrinsic. Let $r(t)$ be a ray, parameterized by its starting point $P_0 \in R^3$ and direction vector $V \in R^3$, $P(t) = P_0 + tV, t \geq 0$. Let P be a bounded plane, where $N, d$ is the normal, represent its normal and offset,   $P \cdot N + d = 0$

The intersection point $P_i$ of the ray and plane could be solved by substitution: \begin{align}
	(P_0&+ t\overrightarrow{V}) \cdot \overrightarrow{N} + d = 0\\
    t &= -(P_0 \cdot \overrightarrow{N} + d) / (\overrightarrow{V} \cdot \overrightarrow{N})\\
    P_{i} &= P_0 + t\overrightarrow{V}
\end{align}

If there are multiple planes intersecting with the ray, take the plane with smallest $t$. Then we can calculate the distance s$r$ between the intersection point $P_{i}$ and plane center $P_{c}$, and the angle between $\overrightarrow{P_{i} - P_{c}}$ and the primary axis of this plane. Finally, we can calculate the predicted color for this pixel is as: \begin{align}
\overrightarrow{V_{axis}} &= PCA(\text{plane points})[0] \\
r &= \sqrt{||P_{i} - P_{c}||^2} \\
\theta &= arccos(\overrightarrow{P_{i} - P_{c}} \cdot \overrightarrow{V_{axis}})\\
\text{color} &= f_{color}(\theta, r) \cdot \text{sigmod} (f_{boundary}(\theta) - r)
\end{align}

The predicted color depends on $N, d, f_{boundary}, f_{color}$, which could be optimized using MSE loss with regard to input frames.
\begin{figure}[t]
	\centering
	\includegraphics[width=0.9\linewidth]{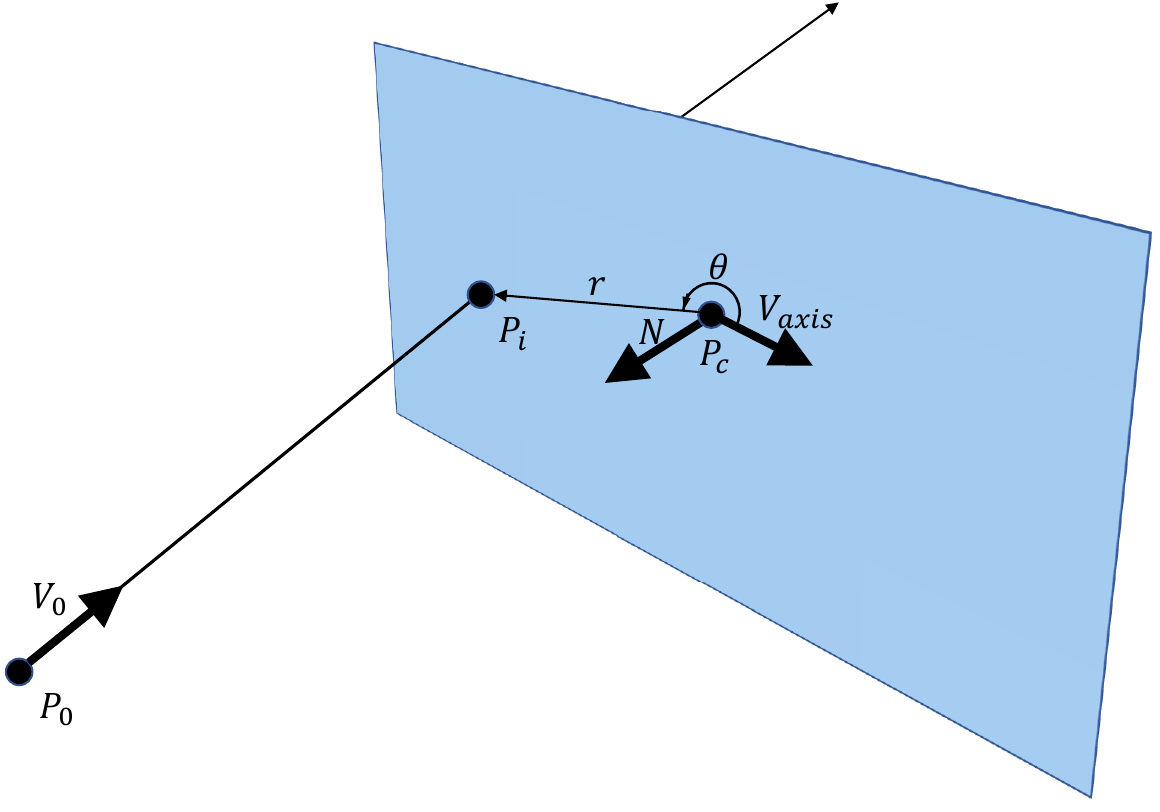}
	\caption{
		\textbf{Refine planes using differentiable rendering}}
	\label{fig:rendering}
\end{figure}
\vspace{1.0cm}

\begin{figure*}[h!]
    \centering
    \vspace{-2em}
       \includegraphics[width=0.95\linewidth]{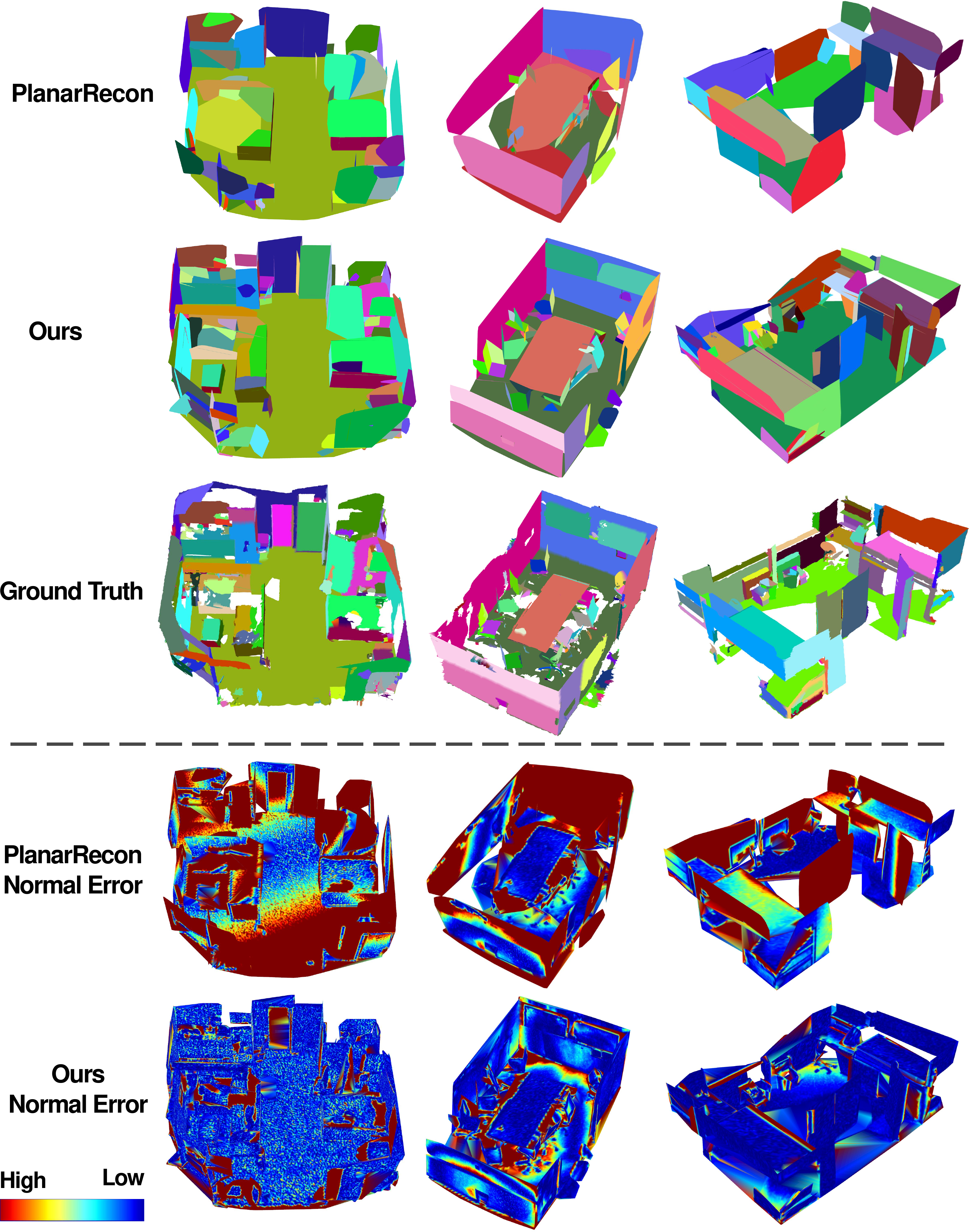}
       \caption{
           \textbf{Qualitative Results for Plane Detection on ScanNet.}
        Our method outperforms PlanarRecon. Our approach is able to reconstruct a more complete scene and retrain more details.
        Different colors indicate different surfaces' segmentation, from much we can observe \OURS{} achieves a much better result in both precision and recalls. The spectrum color indicates surface normal. Compared to PlanarRecon, \OURS{} predicts much lower normal errors.}
       \label{fig:qualitative}
\end{figure*}

\begin{table*}[h!]
	\centering
	\caption{\textbf{3D geometry metrics on ScanNet.} 
		Our method outperforms the compared approaches by a significant margin in almost
		all metrics.
		$\uparrow$ indicates bigger values are better, $\downarrow$ the opposite.
		The best numbers are in bold.
		We use two different validation sets following Atlas \cite{murez2020atlas} (top block) and PlaneAE \cite{yuSingleImagePiecewisePlanar2019} (bottom block).
	}
	\label{tab:scannet-3d}
	\resizebox{1.0\textwidth}{!}{
		\begin{tabular}{cccccc>{\columncolor[gray]{0.902}}ccc}
			\Xhline{3\arrayrulewidth}
			Method             & validation set                               & Comp $\downarrow$ & Acc $\downarrow$      & Recall $\uparrow$         & Prec $\uparrow$           & \textbf{F-score} $\uparrow$  & Max Mem. (GB) $\downarrow$    & Time ($ms$/keyframe) $\downarrow$   \\ \hline
			NeuralRecon~\cite{sun2021neucon} + Seq-RANSAC & \multirow{4}{*}{Atlas~\cite{murez2020atlas}}    & 0.144           & 0.128           & 0.296           & 0.306           & 0.296    & \textbf{4.39}       & 586   \\ 
			Atlas \cite{murez2020atlas} + Seq-RANSAC  &              & 0.102           & 0.190  & 0.316           & 0.348           & 0.331  & 25.91 & 848     \\ 
			ESTDepth \cite{Long_2021_CVPR} + PEAC \cite{feng2014fast}& &  0.174  & 0.135         & 0.289  & 0.335           & 0.304     & 5.44     & 101   \\ 
			PlanarRecon \cite{xie2022planarrecon} &                                                    & 0.154           & \textbf{0.105}   & 0.355           & 0.398  & 0.372 & 4.43 & \textbf{40}  \\
			\textbf{Our} &                                                    & \textbf{0.094}           & 0.133   & \textbf{0.429}           & \textbf{0.409}  & \textbf{0.418} & 8.23 &  44 \\
			\Xhline{3\arrayrulewidth}
			PlaneAE \cite{yuSingleImagePiecewisePlanar2019} & \multirow{2}{*}{PlaneAE~\cite{yuSingleImagePiecewisePlanar2019}}           & 0.128           & 0.151           & 0.330           & 0.262           & 0.290   & 6.29        & \textbf{32}     \\ 
			PlanarRecon \cite{xie2022planarrecon}          &                                         & 0.143  & \textbf{0.098}  & 0.372  & 0.412  & 0.389 & \textbf{4.43} & 40     \\ 
			\textbf{Our}           &                                         & \textbf{0.113}  & 0.126  & \textbf{0.446}  & \textbf{0.415}  & \textbf{0.429} & 8.23 & 44     \\ 
			\Xhline{3\arrayrulewidth}
		\end{tabular}
	}
\end{table*}

\section{Experiments}
This section discusses the implementation details (Sec. \ref{sec:implementation}), the evaluation setup (Sec.~\ref{sec:setup}), quantitative and qualitative evaluations, and ablation studies.
\subsection{Implementation Details}
\label{sec:implementation}B
We use \texttt{torchsparse}~\cite{tang2022torchsparse} to implement the 3D backbone composed of 3D sparse convolutions. The image backbone is a variant of MnasNet \cite{Tan_2019_CVPR} and is initialized with the weights pre-trained from ImageNet~\cite{deng2009imagenet}.
The entire network is trained end-to-end with randomly initialized weights except for the image backbone. The occupancy score $o$ is predicted with a Sigmoid layer. The voxel size of the last level is $4cm$. The number of detection queries is 100.

\subsection{Setup}
\label{sec:setup}
\PAR{Datasets.} 
We perform the experiments on ScanNetv2~\cite{dai2017scannet}. 
The ScanNetv2 dataset contains 1613 RGB-D video sequences taken from indoor scenes by a mobile device mounted by a depth camera. The camera pose is associated with each frame.
As no ground truths are provided in test set, we follow PlaneRCNN~\cite{liu2019planercnn} to generate 3D plane labels on the training and validation sets. Our method is evaluated on two different validation sets with different scene splits used in previous works~\cite{murez2020atlas, liu2018planenet}.  

\PAR{Evaluation Metrics.}
We evaluate the performance of our method in terms of the 3D plane detection, which can be evaluated using instance segmentation and 3D reconstruction metrics following previous works. For plane instance segmentation, due to the geometry difference between the ground truth and prediction meshes, we follow the semantic evaluation method proposed in~\cite{murez2020atlas}. 
More specifically, given a vertex in the ground truth mesh, we first locate its nearest neighbor in the predicted mesh and then transfer its prediction label. 
We employed three commonly used single-view plane segmentation metrics~\cite{yangRecovering3DPlanes2018,liu2019planercnn, tanPlaneTRStructureGuidedTransformers2021,yuSingleImagePiecewisePlanar2019} for our evaluation: rand index (RI), variation of information (VOI), and segmentation covering (SC).
We also evaluate the geometry difference between predicted planes and ground truth planes. 
We densely sample points on the predicted planes and evaluate the 3D reconstruction quality using 3D geometry metrics presented by Murez \etal~\cite{murez2020atlas}. 

\PAR{Baselines.}
Since there are no previous work that focus on learning-based multi-view 3D plane detection, we compare our method with the following three types of approaches:
(1) single-view plane recovering~\cite{yuSingleImagePiecewisePlanar2019}; 
(2) multi-view depth estimation~\cite{Long_2021_CVPR} + depth-based plane detection~\cite{feng2014fast}; and
(3) volume-based 3D reconstruction \cite{sun2021neucon, murez2020atlas} + Sequential RANSAC~\cite{10.1145/358669.358692}.
(4) PlanarRecon~\cite{xie2022planarrecon}

Since baselines (1) and (2) predict planes for each view, we add a simple tracking module to merge planes predicted by the baseline in order to provide a fair comparison. 
The tracking and merging process we designed for our baselines is detailed in the supplementary material.
We use the same key frames as in PlanarRecon for baselines (1) and (2).
For (3), we first employ~\cite{sun2021neucon, murez2020atlas} to estimate the 3D mesh of the scene, and perform sequential RANSAC to group the oriented vertices of the mesh into planes. 
Please refer to our supplementary material for the details of the sequential RANSAC algorithm.
For \cite{sun2021neucon, murez2020atlas}, we run sequential RANSAC every time when a new 3D reconstruction is completed to achieve incremental 3D plane detection.

\subsection{3D Geometric Metric}
We first use RANSAC~\cite{10.1145/358669.358692} method to extract planes from dense mesh, and then we use the extracted planes as the ground truth targets/supervision to do plane detection. A qualitative result is available showed in Figure~\ref{fig:qualitative}. In the plane-only setting, our method correctly reconstruct wall corners without confusion on the wall offsets, because our method explicitly reconstructs height from each surface.
With the texture map and height map enabled, our method is able to reconstruct the scene in high fidelity.

\begin{table}[h!]
	\small
	
	\centering
	\caption{\textbf{3D plane segmentation metrics on ScanNet.} 
		Our method also outperforms the competing baseline approaches in almost all metrics when evaluating plane segmentation metrics.
		$\uparrow$ indicates bigger values are better, and $\downarrow$ indicates the opposite, namely smaller values are better.
		The best numbers are in bold. We use two different validation sets following previous work Atlas \cite{murez2020atlas} (top block) and  PlaneAE \cite{yuSingleImagePiecewisePlanar2019} (bottom block).
	}
	\label{tab:segmt_eval}
	
	\begin{tabular}{ccccc}
		\Xhline{3\arrayrulewidth}
		Method                                              & VOI $\downarrow$                    & RI $\uparrow$      & SC $\uparrow$    \\
		\hline
		NeuralRecon \cite{sun2021neucon} + Seq-RANSAC  & 8.087                               & 0.828                & 0.066               \\
		Atlas \cite{murez2020atlas} + Seq-RANSAC            & 8.485                               & 0.838                & 0.057      \\
		ESTDepth \cite{Long_2021_CVPR} + PEAC \cite{feng2014fast} & 4.470                          & 0.877                & 0.163                 \\ 
		PlanarRecon                                                & 3.622                      & 0.897       & 0.248               \\
		Our                                                & \textbf{3.215}                      & \textbf{0.905}       & \textbf{0.288}               \\
		\Xhline{3\arrayrulewidth}
		PlaneAE \cite{yuSingleImagePiecewisePlanar2019}        & 4.103                               & \textbf{0.908}                 & 0.188              \\
		PlanarRecon                                                & 3.622                      & 0.898      & 0.247     \\
		Our                                                & \textbf{3.210}                      & 0.905      & \textbf{0.288}     \\
		\Xhline{3\arrayrulewidth}
	\end{tabular} 
\end{table}

\subsection{3D Segmentation Metric}
We also perform quantitative evaluation for plane detection using 3D metric, the result is available in Table~\ref{tab:segmt_eval}. Our method produces a more complete and detailed reconstruction because it has extracted major planes in the scenes and using rendering loss to perform pixel level optimization.

\subsection{Qualitative Results}
We provide the qualitative results in Figure~\ref{fig:qualitative}. Visually, we compare with our baseline method PlanarRecon in terms of reconstructed plane surfaces as well as the normal errors. We visualize the reconstructed surfaces and indicate the them by different colors. From Figure~\ref{fig:qualitative}, we can easily observe that our approach \OURS{} achieve much better visual results, \ie, the plane instances are more precisely reconstructed, and more plane instances are reconstructed. The visual results matche the quantatative results in terms of precisions and recalls. We also visualize the heat maps of normal errors. In Figure~\ref{fig:qualitative}, red colors indicate higher errors while blue color means less normal errors. To this end, it is easy to observe that \OURS{} has less normal errors in the visualizations.

\begin{table}[b]
    \centering
    \caption{\textbf{Compare adding standard deviation in feature volume} 
    $\uparrow$ indicates bigger values are better, $\downarrow$ the opposite.
    The best numbers are in bold.
    }
    \label{tab:ablation_view}
   
    \begin{tabular}{cccc>{\columncolor[gray]{0.902}}ccc}
    \Xhline{3\arrayrulewidth}
    Method              & Recall $\uparrow$         & Prec $\uparrow$           & F1-score $\uparrow$  \\ \hline
    PlanarRecon             & 0.355           & 0.398  & 0.372 \\
    PlanarRecon w/ std  &           0.359          & \textbf{0.412}  & 0.383 \\
    UniPlane w/o std &           0.404          & 0.387  & 0.395 \\
    UniPlane &           \textbf{0.429}          & 0.409  & \textbf{0.418} \\
    \Xhline{3\arrayrulewidth}
    \end{tabular}
\end{table}

\subsection{Ablation Study}
In this section, we ablate our designs of choices in our method, including adding standard deviation in feature volume, using object query, and the effectiveness of refine surface with differentiable rendering.

\subsubsection{Adding standard deviation between views in feature volume}
We would like to verify whether adding standard deviation between views is helpful. We compare 4 models: (1) Standard PlanarRecon (2) PlanarRecon with std between views (3) Proposed UniPlane, without std between views (4) Proposed UniPlane, full model.

We present the experiment results in Table. \ref{tab:ablation_view}. Experiment results show that adding standard deviation between views in feature volume increases F1 score for PlanarRecon and \OURS{} model by 0.011 and 0.023, respectively. The improvements on \OURS{} is more significant, we conclude the query based plane detection method could utilize these information more effectively. 

\begin{table}[t]
	\small
	\centering
	\caption{\textbf{Compare query vs heuristic tracking} 
	}
	\label{tab:ablation_tracking}
	
	\begin{tabular}{cccc>{\columncolor[gray]{0.902}}ccc}
		\Xhline{3\arrayrulewidth}
		Method              & Recall $\uparrow$         & Prec $\uparrow$           & F1-score $\uparrow$  \\ \hline
		UniPlane w/heuristic tracking  &           0.377          & 0.398  & 0.387 \\
		UniPlane &           \textbf{0.429}          & \textbf{0.409}  & \textbf{0.418} \\
		\Xhline{3\arrayrulewidth}
	\end{tabular}
\end{table}

\subsubsection{Query vs heuristic tracking}
We would like to verify whether using plane queries from previous fragment to track planes is more effective than heuristic based tracking. We compare 2 models: (1) Proposed \OURS{}, with heuristic tracking (2) Proposed UniPlane, full model. PlanarRecon can not be used with query-based tracking, as there is no plane queries learned.

We present the experiment results in Table. \ref{tab:ablation_tracking}. We observed both recall and precision is lower with heuristic tracking, while recall is decreased more significantly. We conclude its due to heuristic tracking tend to incorrectly match nearby planes having similar plane normal. We can observe this in Figure \ref{fig:qualitative} that heuristic tracking method is incorrectly fusing nearby planes.


\section{Conclusion and Future work}
In this paper, we present a novel method \OURS{} that reconstructs any 3D scenes from a posed monocular video. We have showed its outstanding performance over current state-of-the-art methods on plane detection, and object reconstruction. Our method could be generalized as reconstruct 3D objects using primitive geometries and associated texture and height map. So far we only use plane as the primitive geometry, and in the future work we plan to extend this work to use box, sphere, or Non-Uniform Rational B-Splines (NURBS) \cite{PiegTill96} surface as the primitive, which gives the model much better expressiveness while still keeping the representation compact and interpretable ability.

\pagebreak

{\small
\bibliographystyle{ieee_fullname}
\bibliography{egbib}
}

\end{document}